\documentclass[sigconf]{acmart}

\AtBeginDocument{%
  \providecommand\BibTeX{{%
    \normalfont B\kern-0.5em{\scshape i\kern-0.25em b}\kern-0.8em\TeX}}}

\setcopyright{none}
\renewcommand\footnotetextcopyrightpermission[1]{} 
\usepackage{caption}
\usepackage{subcaption}
\usepackage{graphicx}

\usepackage{amsmath}
\newcommand{\rfme}{\mathtt{RE\mbox{-}RFME}}

\begin{document}

\title{$\rfme$: Real-Estate RFME Model for customer segmentation}


\author{Anurag Kumar Pandey}
\affiliation{%
  \institution{Housing.com}
  \city{Gurgaon}
  \country{India}}
\email{anurag.pandey@housing.com}

\author{Anil Goyal}
\affiliation{%
  \institution{Housing.com}
  \city{Gurgaon}
  \country{India}}
\email{anil.goyal@housing.com}

\author{Nikhil Sikka}
\affiliation{%
  \institution{Housing.com}
  \city{Gurgaon}
  \country{India}}
\email{nikhil.sikka@housing.com}

\renewcommand{\shortauthors}{Pandey, et al.}

\begin{abstract}
Marketing is one of the high-cost activities for any online platform. 
With the increase in the number of customers, it is crucial to understand customers based on their dynamic behaviors  to design effective marketing strategies.
Customer segmentation is a widely used approach to group customers into different categories and design the marketing strategy targeting each group individually.
Therefore, in this paper, we propose an end-to-end pipeline $\rfme$ for segmenting customers into 4 groups: \textit{high value}, \textit{promising}, \textit{need attention}, and \textit{need activation}.
Concretely, we propose a novel RFME (\textbf{R}ecency, \textbf{F}requency, \textbf{M}onetary and \textbf{E}ngagement) model to track behavioral features of customers and segment them into different categories.
Finally, we train the K-means clustering algorithm to cluster the user into one of the $4$ categories.
We show the effectiveness of the proposed approach on real-world Housing.com datasets for both website and mobile application users.
\end{abstract}




\maketitle

\section{Introduction}
Over the past few years, the demand for online real-estate tools has increased drastically due to the ease of access to the internet, especially in developing countries like India. 
There are many online real-estate platforms (e.g., Housing.com, Proptiger.com, Makaan.com, etc.) for owners, developers, and real-estate brokers to post properties for buying and renting purposes.
Daily, these platforms have $115$k and $73$k  customers coming on website and mobile application respectively. 
Given the number of customers, it is crucial to better understand these customers based on static demographics and dynamic behaviors for various downstream tasks such as marketing activities\cite{kotler2010principles, sohrabi2007customer}.
Different marketing activities are deployed to increase customer satisfaction and loyalty.
However, a common marketing strategy for all customers becomes less effective when customer behavior becomes more complicated and changes over time. 
Customer segmentation is a widely used approach to group customers into different categories and marketing strategy can target each group individually\cite{wu2005research,hughes2005strategic,cheng2009classifying,wei2010review}.
In literature, multiple studies have been conducted to understand different variables for the market segmentation  of customers. 
\textit{Wu et al.} \cite{wu2009integrated} and \textit{Kottler et al.} \cite{kottler2009marketing}, showed that customers can be categorized based on their customer characteristics and behavioral variables.
Specifically, customer characteristics include geographic, demographic and psychographic variables, whereas behavioral variables includes the response  of customers toward any product, situation or brand.
\textit{Hughes} \cite{hughes2005strategic, hughes1996boosting} proposed a RFM model which classifies customers into different groups based on three variables: \textbf{R}ecency, \textbf{F}requency and \textbf{M}oney. 
Here, recency refers to the time interval between the last consumer purchase and the present; frequency denotes the number of transactions within a specified time interval; and money represents the amount of money spent within  specific time period.
Businesses that don't have monetary variables, could use engagement variables instead of monetary factors such as bounce rate, visit duration, number of pages visited, time spent per session, etc. This leads to a variation of RFM model i.e. RFE model. 

In past, based on RFM or RFE variables, multiple studies \cite{anitha2019rfm,monalisa2019analysis,khajvand2011estimating, griva2018retail} have been conducted to segment the customers using different clustering algorithms such as K-Means \cite{sculley2010web}, DBSCAN\cite{ester1996density}, etc. 
In this paper, we propose a pipeline (called as $\rfme$) based on novel RFME (\textbf{R}ecency, \textbf{F}requency, \textbf{M}onetary and \textbf{E}ngagement) model for segmenting real-estate customers into different categories for various marketing activities. 
Concretely, we considered the following set of features for recency, frequency, monetary and engagement behaviors for any customer:
\begin{itemize}
    \item Last visit date is considered as indicative of \textbf{recency}.
    \item Total number of sessions in last $45$ days is considered as feature for \textbf{frequency}. Please note that a session is a set of user interactions on the platform within a given time frame. A session can have multiple page views, events, social interactions, and leads.
    \item The total number of product detail pages visited and a total number of dropped leads  are indicative of \textbf{monetary} behavior.  Here, we count all the product detail page views and leads dropped within all the sessions during the last $45$ days.
    \item For \textbf{engagement score}, we track the sessions over the last $45$ days where users performed activities such as applied filters, shortlisted properties, opened customer relationship forms, product detail page views, and dropped leads.

\end{itemize}

\begin{figure*}[!htb]
      \includegraphics[scale=0.38]{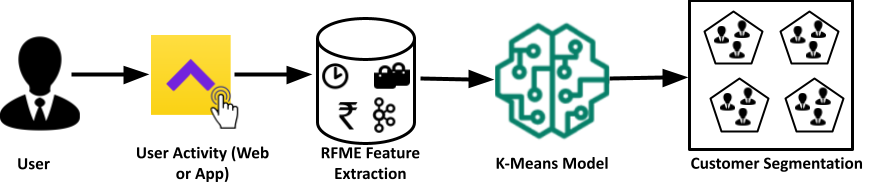}
      \caption{An End-to-end pipeline for proposed  $\rfme$ system}
      \label{fig:flow_diagram}
\end{figure*}

Finally, using above RFME variables (in other words, features) we train the K-Means clustering algorithm to segment the customers into different groups. 
We experimentally show the efficiency of proposed approach on real-world datasets obtained from Housing.com's website and mobile application. 
We are able to segment customers into 4 groups: \textit{high value}, \textit{promising}, \textit{needs attention}, and \textit{needs activation}.
Based on identified groups, we are able to target each group individually using a customized marketing strategy. 

In the next section, we present the proposed RFME model along with the complete pipeline followed by experimental results in section \ref{sec:results}.

\section{The proposed $\rfme$ Pipeline}
\label{sec:algorithm}

In this section, we present an end-to-end pipeline of the proposed system. 
As shown in figure \ref{fig:flow_diagram}, once the user lands on the Housing.com platform (via website or application), he/she searches for properties in a particular locality of the city with an option to apply various preference filters such as bedroom count, age of the property, property type, etc. 
The user explores the recommended properties on the search result page and navigates to the specific product details page of the property in which the user is interested.
The user can express interest by shortlisting the interested properties by marking them favorites or by dropping a lead to the seller/owner of the property using the CRF (customer relationship form) form. 
We use Google Analytics to track the user interaction events on our platform. 
For various marketing activities, it is important to categorize customers based on their geographic, demographic, psychographic and behavioral variables toward any product, situation or brand\cite{wu2009integrated, kottler2009marketing, hughes2005strategic, hughes1996boosting}.
Therefore, in this paper, we propose a novel RFME (\textbf{R}ecency, \textbf{F}requency, \textbf{M}onetary and \textbf{E}ngagement) based model for segmenting real-estate customers into different categories based on tracked events on the website or mobile application.
Under RFME analysis, each customer is scored based on the four following factors:
\begin{itemize}
    \item \textit{Recency}: It refers to the number of days before the current date when a customer made the last visit to the platform. Lesser the value of recency, the higher the customer's visit to the platform.
    \item \textit{Frequency}: It is the total number of sessions (in other words, visits) of customer over the last $45$ days. Please note that a session is a set of user interactions on the platform within a given time frame. A session can have multiple page views, events, social interactions, and leads. The higher the value of frequency, the more the customer visits the platform.
    \item \textit{Monetary}: It is the weighted sum of total number of product detail pages (PDP) visited and the total number of dropped leads  over the last $45$ days which is defined as follows:
    $$\mathtt{Monetary} = ({\mathtt{PDP}})*1 + ({\mathtt{Leads}})*7$$
    In the above equation, we can see that more weightage is given to leads dropped as compared to product detail page visits. This is based on our analysis that a user visits $7$ product detail pages to drop $1$ lead. Please note that a user doesn't buy/rent a house on Housing.com's platform.
    Typically, it takes 3-6 months for a user to finalize any house for buy/rent. 
    Housing.com charges fees from owners/developers/brokers for listing their properties.
    In return, Housing.com provides leads to the owners/developers/brokers. 
    Therefore, we treat the number of leads dropped by a user as monetary behavior.
    \item \textit{Engagement}: It is the sum of the sessions over the last $45$ days where the user performed various activities and it is defined as follows:
    \begin{equation*}
        \begin{aligned}
        \mathtt{Engagement} =  & \mathtt{Filters}_{\mathtt{session}} + {\mathtt{PDP}}_{\mathtt{session}} + \mathtt{Leads}_{\mathtt{session}} \\ 
        & + \mathtt{CRF}_{\mathtt{session}} + \mathtt{Shortlistings}_{\mathtt{session}} 
        \end{aligned}
    \end{equation*}
    where, $\mathtt{Filters}_{\mathtt{session}}, {\mathtt{PDP}}_{\mathtt{session}}, \mathtt{Leads}_{\mathtt{session}}, \mathtt{CRF}_{\mathtt{session}},$ and $\mathtt{Shortlistings}_{\mathtt{session}}$ are the number of sessions where user has applied some filters while search properties, has visited PDP page, dropped leads,  opened customer relationship form (CRF), and shortlisted properties respectively. 
    Compared to the monetary score, here we track user engagement at the session level. 
    In the case of the monetary score, it is important to track all the dropped leads within sessions. 
    However, in engagement score we track the overall user activities on the platform.
\end{itemize}

Finally, based on above calculated RFME scores for each user over last $45$ days, we train the K-Means clustering algorithm to segment customers into categories. 

\section{Experiments}
\label{sec:results}
In this section, we present the obtained experimental results using the proposed pipeline.
\subsection{Datasets:} 
We have collected the historical user event data separately for both web and mobile application from the Housing.com's database from   24 December 2022 to 22 January 2023.  
We have considered first $20$ days of data as training data and last $10$ days as testing data. 
The general statistics for both web and application dataset is presented in Table \ref{tab:Web & App train test data}.


\begin{table}[]
\begin{tabular}{|l|l|l|l|}
\hline
\multicolumn{1}{|c|}{\textbf{S.No.}} & \multicolumn{1}{c|}{\textbf{Period}} & \multicolumn{1}{c|}{\textbf{Web}} & \multicolumn{1}{c|}{\textbf{App}} \\ \hline
$\mathtt{Train}$                    & $\mathtt{24 \ Dec,22 - 11 \ Jan,23}$          & $2,056,896$                           & $392,375$                            \\ \hline
$\mathtt{Test}$                        & $\mathtt{12 \ Jan,23 - 22 \ Jan,23}$              & $788,946$                            & $352,457$                            \\ \hline
\end{tabular}

\caption{Train and Test Data Split for Web and App}
\label{tab:Web & App train test data}
\end{table}

\subsection{Experimental Results}
To find the clusters for customer segmentation, we used the distance-based \textit{K-means} unsupervised clustering algorithm where data points that are close to each other are grouped in a given number of clusters/groups.
Before training the K-means algorithm, it is essential to identify the number of clusters in any given dataset.
We used \textit{elbow method} to identify the number of clusters that calculates the $\mathtt{WCSS}$ (Within-Cluster Sum of Square) i.e. the sum of the square distance between points in a cluster and the cluster centroid for a different number of clusters ($\mathtt{k}$). 
In elbow method, we plot the $\mathtt{WCSS}$ versus $\mathtt{k}$ graph and we select $\mathtt{K}$ value where increasing the value of ‘$\mathtt{K}$’ does not lead to a significant reduction in WCSS.
For both elbow method and K-mean clustering algorithm, we used the implementation of $\mathtt{scikit\mbox{-}learn}$ python package with default hyperparameters.



\subsubsection{Optimal number of Clusters:}
\begin{figure*}
\centering
\begin{subfigure}{.5\textwidth}
  \centering
  \includegraphics[scale=0.35]{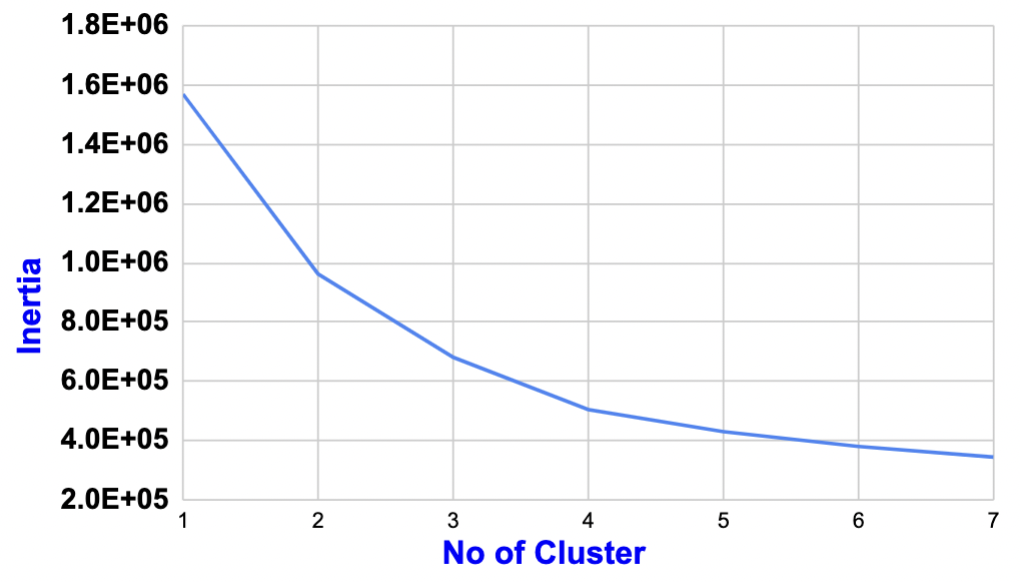}
  \caption{App Users}
  \label{fig:number_clusters_web}
\end{subfigure}%
\begin{subfigure}{.5\textwidth}
  \centering
  \includegraphics[scale=0.33]{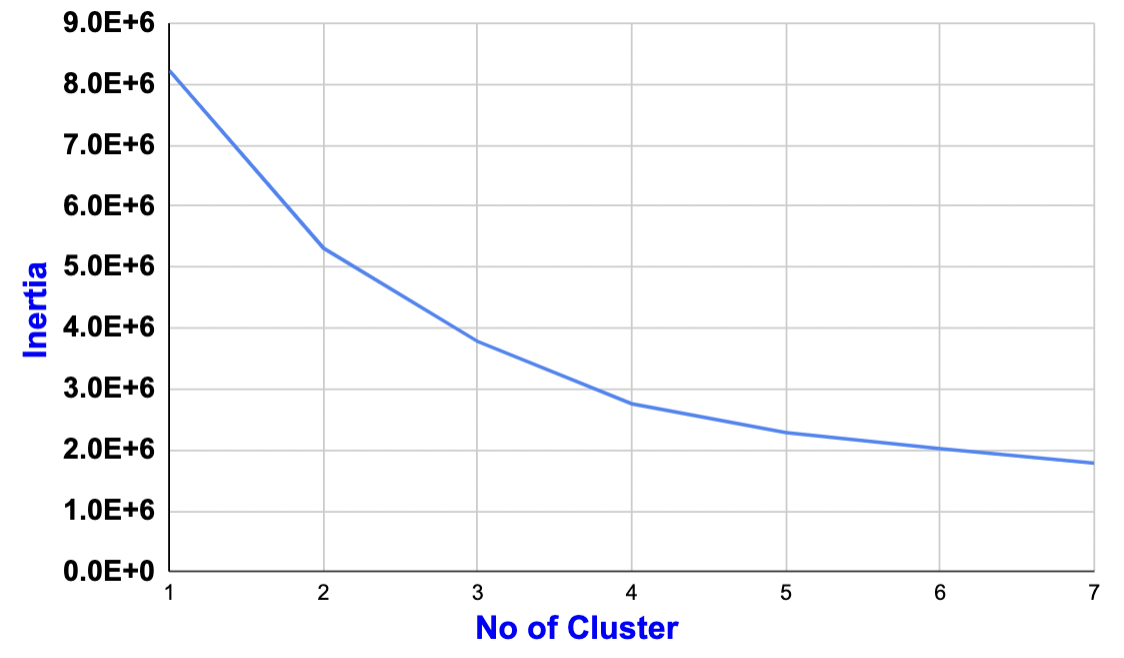}
  \caption{Web Users}
  \label{fig:number_clusters_web}
\end{subfigure}
\caption{Number of Clusters for App and Web}
\label{fig:number_clusters}
\end{figure*}

We plotted the $\mathtt{WCSS}$ versus $\mathtt{k}$ graph (both for web and app users) using the elbow method for different values of $\mathtt{k}$ from $1$ to $7$. 
From Figure \ref{fig:number_clusters}, we can easily deduce that after $\mathtt{k}=4$ (for both app and web users) increasing the value of $\mathtt{k}$ doesn't lead to significant decrease in WCSS score. 
Therefore, we picked $\mathtt{k}=4$ as optimal number of clusters for both web and app users.


\subsubsection{Cluster Definition \& Characteristics:}
We have trained the K-means algorithm on both web and app datasets using $\mathtt{k}=4$ clusters. 
On training datesets, the mean statistics for all the $4$ identified clusters are presented in tables \ref{tab:App Cluster Characteristics} and \ref{tab:Web Cluster Characteristics}.
From the tables, we are able to classify customers into the following four segments:
\begin{itemize}
    \item \textbf{High Value}: Customers in this segment have ideal behavior with very high frequency, monetary and engagement values compared to any other cluster. 
    In order to increase the recency, we can recommend them similar properties for which they have shown interest in past. 

    \item \textbf{Promising}: Customers in this segment have similar recency as compared to \textit{high value} customers but lower FME scores. 
    For these customers, the marketing team needs to upsell the properties by  recommending similar properties over email or mobile message.
    
    \item \textbf{Needs Attention}: Customers in this segment have high recency (more recent on the platform) but low FME scores. 
    These customers are probably in the exploration phase and looking for properties on multiple platforms.
    Therefore, the marketing team needs to give special attention to these customers possibly by reaching out to them over the phone to understand their needs or giving them offers.

    \item \textbf{Needs Activation}: Customers in this segment have low RFME scores as compared to any other cluster.
    Here, customers are ``less valued'' users who need to be engaged more by creating awareness about the platform through advertisements etc.

\end{itemize}

\begin{table}[]
\resizebox{\columnwidth}{!}{%
\begin{tabular}{|c|c|c|c|c|}
\hline
\textbf{Cluster} &
  \textbf{\begin{tabular}[c]{@{}c@{}}Needs \\ Activation\end{tabular}} &
  \textbf{\begin{tabular}[c]{@{}c@{}}Needs \\ Attention\end{tabular}} &
  \textbf{Promising} &
  \textbf{\begin{tabular}[c]{@{}c@{}}High\\  Value\end{tabular}} \\ \hline
\textbf{Recency}      & 29 & 19 & 23 & 24  \\ \hline
\textbf{Frequency}    & 3  & 4  & 20 & 57  \\ \hline
\textbf{Monetary} & 7  & 7  & 73 & 242 \\ \hline
\textbf{Engagement}   & 3  & 3  & 25 & 77  \\ \hline
\end{tabular}%
}
\caption{Mean Statistics for App training dataset}
\label{tab:App Cluster Characteristics}
\end{table}
\begin{table}[]
\resizebox{\columnwidth}{!}{%
\begin{tabular}{|c|c|c|c|c|}
\hline
\textbf{Cluster} &
  \textbf{\begin{tabular}[c]{@{}c@{}}Needs \\ Activation\end{tabular}} &
  \textbf{\begin{tabular}[c]{@{}c@{}}Needs \\ Attention\end{tabular}} &
  \textbf{Promising} &
  \textbf{\begin{tabular}[c]{@{}c@{}}High\\  Value\end{tabular}} \\ \hline
\textbf{Recency}      & 26 & 16 & 21 & 21  \\ \hline
\textbf{Frequency}    & 1  & 2  & 11 & 41  \\ \hline
\textbf{Monetary} & 2  & 3  & 31 & 123 \\ \hline
\textbf{Engagement}   & 2  & 3  & 21 & 75  \\ \hline
\end{tabular}%
}
\caption{Mean Statistics for Web training dataset}
\label{tab:Web Cluster Characteristics}
\end{table}

\begin{figure*}
\centering
\begin{subfigure}{.5\textwidth}
  \centering
  \includegraphics[scale=0.35]{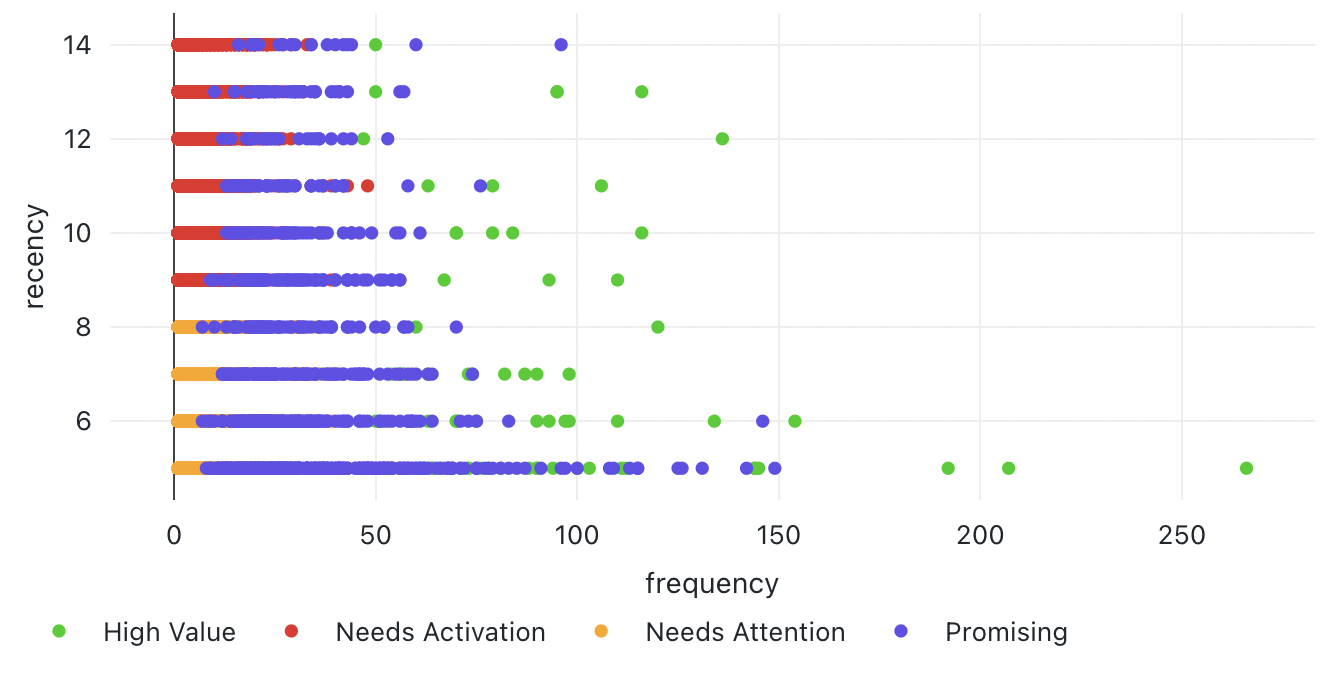}
  \caption{Recency vs Frequency analysis for app users}
  \label{fig:rec_freq_app}
\end{subfigure}%
\begin{subfigure}{.5\textwidth}
  \centering
  \includegraphics[scale=0.35]{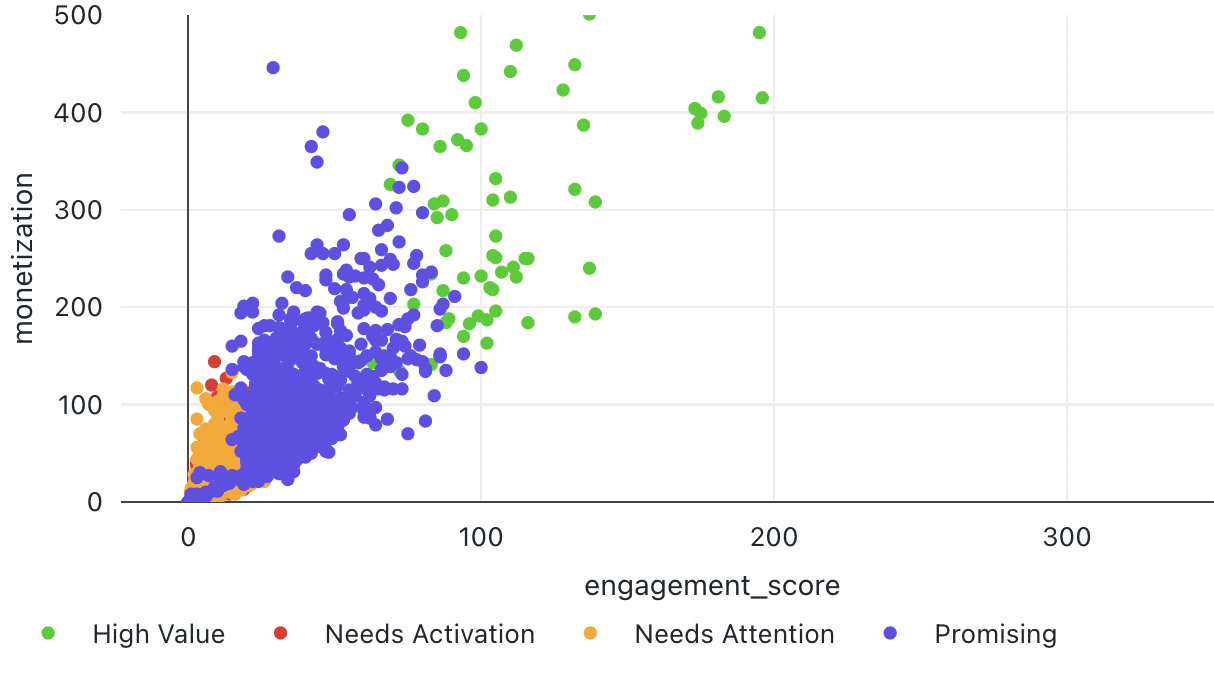}
  \caption{Monetary vs Engagement analysis for app users}
  \label{fig:mon_eng_app}
\end{subfigure}
\caption{Cluster analysis for mobile application test data users}
\label{fig:cluster_analysis_app}
\end{figure*}

\begin{figure*}
\centering
\begin{subfigure}{.5\textwidth}
  \centering
  \includegraphics[scale=0.35]{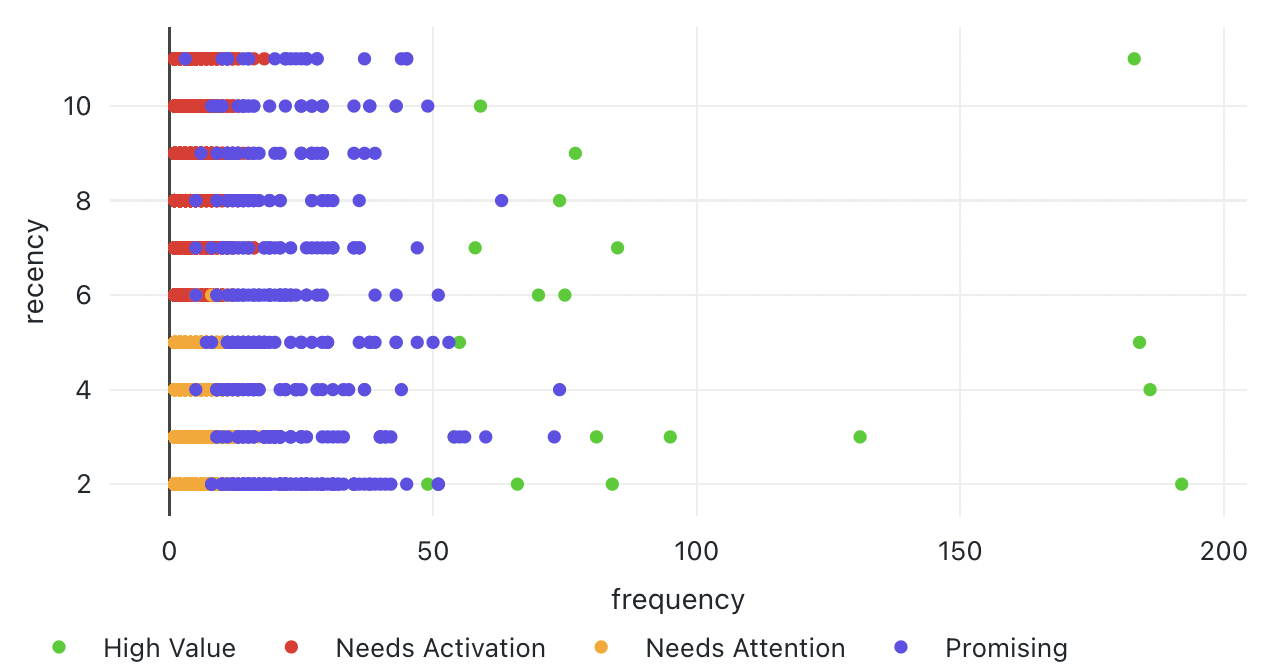}
  \caption{Recency vs Frequency analysis for web Users}
  \label{fig:rec_freq_web}
\end{subfigure}%
\begin{subfigure}{.5\textwidth}
  \centering
  \includegraphics[scale=0.35]{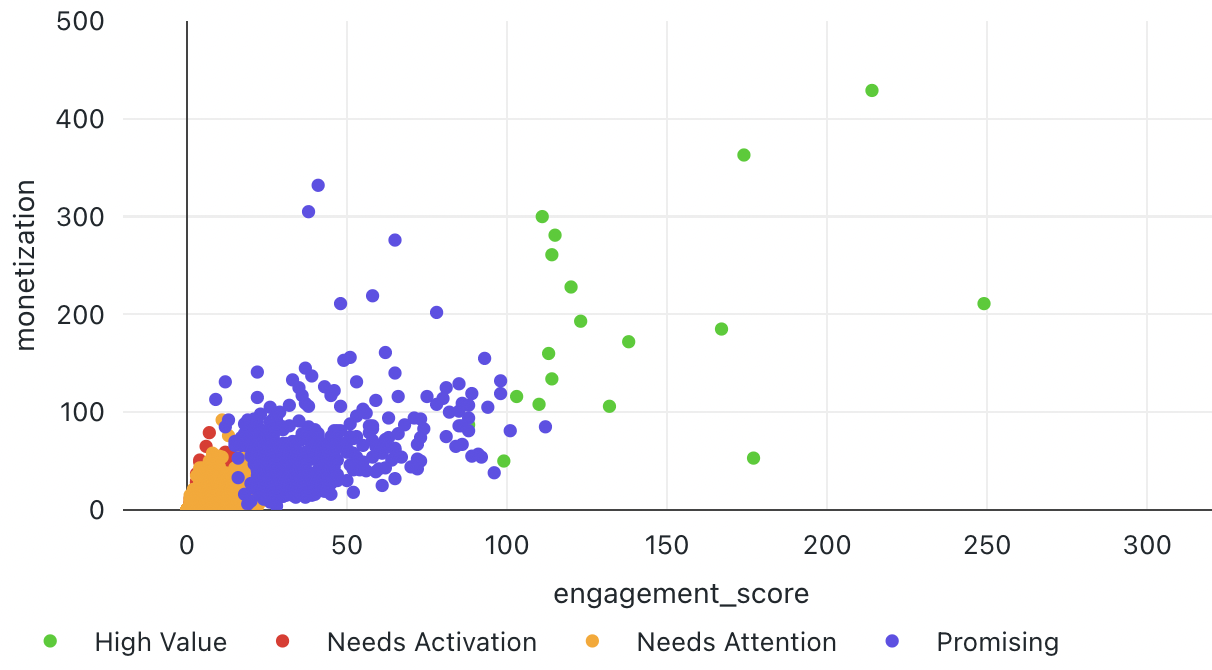}
  \caption{Monetary vs Engagement analysis for web users}
  \label{fig:mon_eng_web}
\end{subfigure}
\caption{Cluster analysis for website application test data users}
\label{fig:cluster_analysis_web}
\end{figure*}

Similar to training dataset results, we have evaluated the above algorithm on test data for both web and mobile application users.
As shown in Figures \ref{fig:cluster_analysis_app} and \ref{fig:cluster_analysis_web}, we plotted the recency vs frequency and monetary vs engagement graphs for app and web users respectively. 
From recency vs frequency graphs, we can deduce that \textit{high value} and \textit{promising} customers have recency ranging from $0$ to $14$ days. However, \textit{high value}    customers have higher frequency as compared to \textit{promising customers}.
Moreover, \textit{Need Attention} are more recent on the platform as compared to \textit{need activation} customers even though they both have low frequency. 
From monetary vs engagement graphs, we can deduce that \textit{high value} customers have higher monetary and engagement scores compared to \textit{promising} customers. 
However, \textit{needs activation} and \textit{needs attention} customers have similar behavior in terms of monetary and engagement scores.
From our experiments, we can deduce that in the real-world production environment the proposed pipeline $\rfme$ is able to correctly classify customers into one of the four segments which can help the marketing teams to take informed decisions while creating marketing strategies or campaigns.




\section{Conclusion}
This paper introduces the $\rfme$ pipeline for customer segmentation of real-estate customers for various marketing activities.
We propose a novel RFME (\textbf{R}ecency, \textbf{F}requency, \textbf{M}onetary and \textbf{E}ngagement) model to group customers into 4 categories: \textit{high value}, \textit{promising}, \textit{need attention}, and \textit{need activation}.
Finally, using RFME features, we train the K-means clustering algorithm and showed the efficacy of the proposed pipeline on two real-world datasets.

\bibliographystyle{ACM-Reference-Format}
\bibliography{bibliography}


\end{document}